\pdfoutput=1

\documentclass[sigconf]{acmart}

\AtBeginDocument{%
  }

\usepackage{arydshln}
\usepackage{colortbl}
\usepackage{placeins}
\usepackage{tabularx}
\usepackage{wrapfig}
\usepackage{graphicx}
\usepackage{soul}
\usepackage{caption}

\newcolumntype{L}[1]{>{\hsize=#1\hsize\raggedright\arraybackslash}X}%
\newcolumntype{R}[1]{>{\hsize=#1\hsize\raggedleft\arraybackslash}X}%
\newcolumntype{C}[1]{>{\hsize=#1\hsize\centering\arraybackslash}X}%

\definecolor{darkred}{RGB}{139,0,0}
\definecolor{verylightred}{RGB}{255,243,240}

\copyrightyear{2024}
\acmYear{2024}
\setcopyright{acmlicensed}
\acmConference[SIGIR '24] {Proceedings of the 47th International ACM SIGIR Conference on Research and Development in Information Retrieval}{July 14--18, 2024}{Washington, DC, USA.}
\acmBooktitle{Proceedings of the 47th International ACM SIGIR Conference on Research and Development in Information Retrieval (SIGIR '24), July 14--18, 2024, Washington, DC, USA}
\acmDOI{10.1145/3626772.3661371}
\acmISBN{979-8-4007-0431-4/24/07}

\begin{document}

\title{Question Suggestion for Conversational Shopping Assistants Using Product Metadata}

\author{Nikhita Vedula}
\affiliation{%
  \institution{Amazon, Seattle}
  \country{}
}
\email{veduln@amazon.com}

\author{Oleg Rokhlenko}
\affiliation{%
  \institution{Amazon, Seattle}
  \country{}
}
\email{olegro@amazon.com}

\author{Shervin Malmasi}
\affiliation{%
  \institution{Amazon, Seattle}
  \country{}
}
\email{malmasi@amazon.com}

\renewcommand{\shortauthors}{Vedula et al.}

\begin{abstract}

Digital assistants have become ubiquitous in e-commerce applications, following the recent advancements in Information Retrieval (IR), Natural Language Processing (NLP) and Generative Artificial Intelligence (AI). 
However, customers are often unsure or unaware of how to effectively converse with these assistants to meet their shopping needs. In this work, we emphasize the importance of providing customers a fast, easy to use, and natural way to interact with conversational shopping assistants. 
We propose a framework that employs Large Language Models (LLMs) to automatically generate contextual, useful, answerable, fluent and diverse questions about products, via in-context learning and supervised fine-tuning. Recommending these questions %
to customers as helpful suggestions or hints to both start and continue a conversation can result in a smoother and faster shopping experience with reduced conversation overhead and friction. 
We perform extensive offline evaluations, and discuss in detail about potential customer impact, and the type, length and latency of our generated product questions if incorporated into a real-world shopping assistant.

\end{abstract}




\maketitle

\section{Introduction and Background}
\label{sec:introduction}

Driven by the recent progress in IR, NLP and generative AI, there has been a surge in the deployment of intelligent virtual assistants and chat bots employing both voice and text, in major e-commerce services like Amazon, Walmart and Shopify. 
Shopping assistants can recommend products, summarize reviews and catalog metadata, compare products, and answer questions about product features, orders and deliveries. 
The adoption of such conversational assistants by customers to meet their diverse shopping needs hinges on accessible, convenient and customer-friendly application interfaces \cite{balakrishnan2021conversational}.
To create sustainable and satisfying interactions between customers and shopping assistants, we must help customers adapt to them, %
by expediting learning curves and minimizing the need for customer trial and experimentation.  

One way of achieving this is to provide customers with a useful set of \textit{question suggestions} or \textit{question hints},  
that can guide customers in initiating and advancing their shopping conversations with assistants. These questions can also inspire them to ask the assistant their own questions based on their needs. 
The automatic Question Generation (QG) problem is well studied across various domains~\cite{mulla2023automatic,ghanem-etal-2022-question,zhuang2024toolqa}. 
However, a shopping assistant %
suggesting irrelevant, unanswerable or redundant questions can severely impair customer trust and the overall buying experience.
In this work, we introduce shoppers to the product-based question answering (QA) functionality of shopping assistants. We propose an LLM-based approach leveraging in-context learning (ICL) \cite{wei2022emergent} and supervised fine-tuning (SFT), to \textit{automatically generate question suggestions} grounded in catalog-derived product metadata and buyer reviews.

Further, we identify and propose several shopping-specific criteria that should be satisfied by our generated product questions, to make interactions with shopping assistants more natural, intuitive and optimal for customers (defined in Section~\ref{sec:criteria}). 
We associate and pair each generated question with its corresponding product context information containing the answer to the question, ensuring that our generated questions are \textit{answerable}. Since this context is derived from the catalog or reviews, it is human authored and has the added benefit of avoiding automatic answer generation~\cite{zhuang2024toolqa,wei2022chain,kojima2022large}, which suffers from a risk of hallucination~\cite{rawte2023survey,zhang2023siren}. 
Questions paired with their relevant answers or context can also be useful in downstream applications such as creating an automatic bank of frequently asked questions (FAQs)~\cite{bihani2018faqtor,mass2020unsupervised}, building a retrieval index for a retrieval augmented generation (RAG) LLM~\cite{gao2023retrieval}, or fine-tuning a model for e-commerce based QA~\cite{deng2023product}.

Finally, we extensively evaluate our approach, and discuss its potential impact on shopping assistants in Sections~\ref{sec:results} and~\ref{sec:deploy}. We expect to obtain a consistent alignment between the offline and online results and performance trends, as well as several valuable insights about real customer interactions with shopping assistants. Providing automatic question suggestions to customers that are a good mix of both broad and specific questions can effectively streamline the conversation, which can in turn lead to low customer dissatisfaction rate. 
The latency of real-time question generation by LLMs can be reduced by mechanisms such as caching specific outputs or delivering the generated tokens in a streaming fashion~\cite{xiao2023efficient}. 
While shorter questions are easier to display (especially on mobile device interfaces), longer questions can serve as useful summaries for multiple key product aspects at a time, saving customers from clicking or typing out a series of questions. 

Overall, our proposed approach reduces the number of steps required for a customer to reach their shopping goal, contributing to an improved overall shopping experience, fostering customer satisfaction and engagement. 
We show examples of questions generated by our approach and their paired answers in Table~\ref{tab:examples}.

\section{Problem Statement and Dataset}
\label{sec:problem}

We formally describe our problem of generating \textit{conversational product-based question suggestions}. 
Every product in an e-commerce catalog is associated with multiple sources of textual content about it, coming from both the product sellers or manufacturers (product descriptions and metadata) as well as buyers of the product (reviews). 
Our objective is to leverage this rich content to create a model capable of generating one or more %
product questions grounded in the input product context, that potential customers may ask shopping assistants about the product.

We use the publicly available Amazon Reviews Dataset~\cite{ni2019justifying} as our information source. It contains at least 200 million buyer reviews for tens of thousands of products spanning more than 20 product categories, 
as well as product catalog information including the title, category, descriptions, price, brand and other product-specific attributes. 
Since not all customer reviews are informative or of high quality, we only use reviews that have received a high number of \textit{helpfulness} votes from other customers, as well as reviews written by reviewers who are a part of Amazon's Vine\footnote{\url{https://www.amazon.com/vine/about}} program.%

\section{Our Approach}
\label{sec:approach}

Our proposed LLM-based question generation approach takes textual content about a product as input (derived from a product review or a catalog description from Section~\ref{sec:problem}), and generates one or more question suggestions for customers from that content. 

\subsection{Product Question Quality Criteria}
\label{sec:criteria}

We first identify and propose a set of crucial, multi-faceted quality criteria required for an effective and efficient conversation between a customer and shopping assistant. Our approach seeks to satisfy these for each generated product question.

(i) \textbf{Relevance:} The question should be applicable and appropriate with respect to the product under consideration and its features.

(ii) \textbf{Usefulness:} The product question (and it's corresponding answer) should provide helpful information to customers, that can benefit them in deciding whether or not to purchase the product.

(iii) \textbf{Answerability:} The answer to the generated product question must be present in its input context (review or catalog snippet).

(iv) \textbf{Fluency:} The generated question should be grammatically correct, fluent,  coherent and easily understandable in general. 

(v) \textbf{Style:} The generated question should mimic a customer's inquiry style. E.g., the question \textit{``Which colors do you prefer for this jacket?"} is about a valid product aspect (\textit{color}), but is framed as a clarification question that an assistant might ask a customer, rather than a valid question that a customer may ask a shopping assistant. A better formulated question suggestion for a customer would be \textit{``Which colors are available for this jacket?"}

(vi) \textbf{Diversity:} The set of questions that are generated per product should cover a wide variety of product features or aspects without being redundant or repetitive. The questions should be of varying lengths (both short and concise as well as long and verbose) and complexities (in terms of their shape and sentence structures), to cater to a multitude of customer requirements. 
Generating and recommending questions of several different types also aids customers at various stages of their shopping journeys. For instance, new customers may prefer broad or exploratory product questions, whereas customers midway during the shopping process may inquire about comparisons or specific product attributes.

We next describe the two LLM-based generation techniques that we leverage as part of our question generation approach.

\begin{table}[t!]
\footnotesize
\begin{center}
\setlength{\arrayrulewidth}{1.5pt}
\rowcolors{1}{verylightred}{verylightred} %
\ttfamily
    \begin{tabular}{|p{0.95\linewidth}| }
    \hline
    \textbf{Human:} You are an intelligent shopping assistant helping customers shop for products. Use the \textbf{Product Info} given below to output the top product question, question type and it's estimated customer interest score (1-10). A customer should be able to ask a salesperson this question.   \\ 
    Your question should be about broad features of the product, specific product aspects, compatibility with other products, comparisons with other products, or other types of important buying guide questions that can fulfill diverse customer needs. \\ \\

    Your question must adhere to the following criteria:

1. Your question must be relevant to the given Product Info \\
2. Your question should be useful in helping the customer decide whether to purchase the given product. \\
3. The answer to your question must be contained within the Product Info. \\
4. Your question should be grammatically correct, fluent and coherent. \\
5. Your question can either be short and concise, or long and verbose based on your judgment of the Product Info. \\
6. Avoid using personal, first person or second person pronouns in your output question. \\
7. Use anaphora like `it' or `this' to refer to the product when applicable. \\ \\

Use the Product Info given below to output the top product question, question type and it's estimated customer interest score (1-10). \\
Output structured pipe-separated columnar data without any other text. \\
\\ 
\textbf{Product Info:} \{data\}. \\
\textbf{Assistant:}
\\ \hline
    \end{tabular}
    \vspace{1.0em}
    \captionof{figure}{Prompt for ICL-based product question generation.}
    \label{tab:prompt}
\end{center}
\end{table}

\subsection{Product Question Suggestion Generation}
\label{sec:ICL_SFT}

\paragraph{\bf In-Context Learning:}
We harness the rich world knowledge possessed by LLMs, coupled with the question quality criteria described in Section~\ref{sec:criteria} to construct a prompt (Figure~\ref{tab:prompt}) that can generate diverse product question suggestions for customers. 
In Section~\ref{sec:results}, we evaluate the performance of this prompt both in a zero-shot manner, and by enriching it with examples that the LLM can learn from in a few-shot manner.

\paragraph{\bf Supervised Fine-tuning:}

We construct a high quality training dataset in a time-efficient manner with reduced human annotation effort, to fine-tune an LLM for our QG task. The dataset consists of parallel pairs of product contexts (snippets of product reviews or catalog descriptions from %
Section~\ref{sec:problem}), and the corresponding questions derived from those contexts.  
Note that the contexts can also be considered as the answers for the generated questions, since our approach by design ensures that the questions generated from the contexts are answerable by the contexts themselves. 

We first collect an initial set of labeled question-context pairs via ICL, for products spanning a diverse set of features and categories. 
We manually inspect these data pairs to ensure that each product question satisfies the criteria described in Section~\ref{sec:criteria}, and if required, clean and/or rephrase the questions. 
This semi-automatic process results in the creation of a high quality supervised dataset of 1.8K question-context pairs. The questions cover a wide variety of product contexts, and range from short, simple questions to long questions specifying multiple aspects. 
We then fine-tune an LLM with this labeled data, to take a product context as input and generate a question about it.

\begin{table}[th!]
  \caption{Evaluating product questions generated by our approach across several quality criteria using GPT-4.}
  \label{tab:autoeval}
  \begin{tabular}{lccl}
    \toprule
    Metric & ICL (zero-shot) & ICL (few-shot) & SFT\\
    \midrule
    Relevance & 0.80 & 0.86 & 0.81 \\
    Usefulness & 0.72 & 0.73 & 0.75 \\
    Answerability & 0.63 & 0.70 & 0.72 \\
    Fluency & 0.98 & 0.99 & 0.95 \\
    Style & 0.61 & 0.66 & 0.70 \\
  \bottomrule
\end{tabular}
\end{table}

\begin{table*}[!t]
\caption{Examples of product contexts and their corresponding question suggestions generated by our proposed approach.}
\label{tab:examples}
\small
\begin{tabularx}{\linewidth}{L{0.2}|L{0.85}|L{0.5}|L{0.35}}
\hline
{\bf Context type} & {\bf Product Context} & {\bf Generated Questions} & {\bf Question Quality} \\
\hline
Review (Commercial toilet paper holder) & UGH! DUMB! I thought this screwed to the wall, not GLUED to the wall! … A great way to destroy a huge portion of my wall and rip it off when I go to remove it or it falls off. And I'm going to assume that if I did peel it off and it didn't take part of my wall/paint with it, that it would leave sticky residue behind…
  & \textbf{Q1:} Does the toilet paper holder stick or screw to the wall, as advertised? \newline
\textbf{Q2:} Will removing the toilet paper holder leave behind sticky residue or cause wall damage?
 &  Both questions are relevant, useful, answerable, fluent, diverse, and summarize the review well.   \\ 
\hline
Review (Cat repellent mats) & Does NOT work for cats - false advertising! …This thing is a joke, and cats are not even afraid of the beep anyway. I wish we could return it for a refund...
  & \textbf{Q1:} Are cats afraid of the beep emitted from the mat?
 &   Relevant, useful, answerable, fluent and concise. \\ 
 \hline
Catalog Description (Rongeurs) & Item Package Height: 20.32 Centimeters. Autoclavable. Medical Grade Steel. Unit Count: 1.0
  & \textbf{Q1:} Can the rongeurs be reused after cleaning and sterilization? 
 &  Relevant, useful, fluent, but unanswerable by the context   \\ 
\hline
 Review (Salted soy nuts) & I loved these because sometimes your looking for a salty snack that doesn't have so much fat in them like potato chips. This product satisfies the urge nicely
  & \textbf{Q1:} Are the salted soy nuts a healthy alternative to potato chips? \newline
\textbf{Q2:} Which snack has more fat, potato chips or the salted soy nuts?
 &  Both questions are relevant, useful, answerable and fluent, but are not very diverse from each other   \\ 
\hline
Catalog Description (Smart doorbell) & Built-in speaker/microphone. Talk to visitors anytime anywhere... Intelligent real-time monitoring via mobile phone... Video recording, picture-taking and screenshot, video playback and storage. Play videos anytime anywhere via mobile phone. When storage is full, the earliest recorded video will be overwritten. 
  & \textbf{Q1:} What are key features of this doorbell? \newline
  \textbf{Q2:} Can the camera enable mobile phone monitoring? \newline
\textbf{Q3:} Can the camera take pictures, record videos, and store them on a mobile device?
 &  Q1, Q2, Q3 are relevant, fluent and answerable. Q1, Q3 are less useful to customers making a purchase decision. Q2, Q3 are not very diverse.  \\ 
 \hline
Review (Hair growth tonic) & This came in a day, very fast. Best ever, was sceptical at first but persevered. I have been using it for like a week and have seen that my eyelashes did look fuller.
  & \textbf{Q1:} How fast was the hair growth tonic delivered, and did it make eyelashes look fuller after use? 
 &  Relevant, fluent, slightly useful, summarizes the review well, answerable.  \\ 
 \hline
\end{tabularx}
\end{table*}

\section{Experiments and Results}

\label{sec:results}

The below design choices prioritize velocity for our offline pilot study, 
in terms of size, speed of training and inference, and performance on a manually curated validation set. However, as discussed in Section~\ref{sec:deploy}, our proposed approach is generalizable to any LLM.

\paragraph{\bf Data:}
We construct a test dataset of 1,000 product contexts, spanning more than 20 product types~\cite{ni2019justifying}, %
that we use to individually evaluate our proposed approach from Section~\ref{sec:approach}.

\paragraph{\bf Models:}
ICL questions are generated by the Claude-2 LLM.\footnote{\url{https://www.anthropic.com/news/claude-2}}  
For SFT, we train the 11B \textsc{Flan-T5-xxl}~\cite{longpre2023flan} model with 1.8K pairs of product contexts and their corresponding question suggestions (from Section~\ref{sec:ICL_SFT}), %
for $8$ epochs with an initial learning rate of $1e^{-5}$. 

\paragraph{\bf Evaluation Metrics:}
We evaluate the questions generated by our approach on each quality dimension defined in Section~\ref{sec:criteria}. %

\paragraph{\bf Automatic Evaluation:}
A diverse set of question suggestions are generated by our approach, with an average diversity $>75\%$ across lists of questions, with respect to question length, lexical shape and product aspects mentioned. 
We use the GPT-4 LLM~\cite{achiam2023gpt} to automatically evaluate the questions generated by our proposed approach, on each dimension described in Section~\ref{sec:criteria}. %
The results are displayed in Table~\ref{tab:autoeval}. 
We observe that both the ICL and SFT variants of our approach achieve good performance on the relevance and fluency dimensions. 
There is a greater scope for improvement in usefulness, answerability and question style, all of which achieve scores higher than $70\%$ with our proposed approach. 
We also notice that more than 55\% of the questions labeled as unanswerable were actually \textit{partially answerable} by the provided context. 
The zero shot and few shot ICL variants perform comparably in case of usefulness and fluency metrics, and sometimes better than the SFT variant (for relevance and style), possibly because Claude is larger and better at following instructions than \textsc{Flan-T5-xxl} \cite{ge2024openagi,zhang2023instruction}.

\paragraph{\bf Human Evaluation:}
We next perform a human evaluation study to examine whether the human judgments made on the quality of the generated product questions align with the automatic evaluation outcomes from GPT-4. We randomly sample and annotate 75 examples from the test set across each quality dimension. %
We observe general trends largely similar to those noted with automatic evaluation. 
We obtain a high overall agreement $>75\%$ between the human and GPT-4 annotations across multiple dimensions. The break down of this agreement for each individual dimension is as follows: 88\% agreement for relevance, 61.33\% for usefulness, 81.33\% for answerability, 90.66\% for fluency, and 66\% for style. We observe a very high agreement between humans' and GPT-4's judgments greater than 80\% for the relatively objective metrics of relevance, fluency and answerability.  
The lower agreement between humans and GPT-4 for usefulness and style is possibly due to some subjectivity involved in interpreting what is a useful or stylistically valid question a customer may ask a shopping assistant. 

\paragraph{\bf Example Question Suggestions:}
Table~\ref{tab:examples} shows diverse examples of product contexts %
and suggestion questions we generated for them. The questions are relevant and fluent, with varied question types and lengths. The longer questions effectively summarize the input contexts (rows 1, 6). Some questions are about product features that are more obvious or less useful (rows 5, 6). Row 3 depicts a question that can be both relevant and useful to potential buyers, but is not derived from the context and cannot be answered by it.

\section{Discussion}
\label{sec:deploy}

Having validated our approach via offline automatic and human evaluations in Section~\ref{sec:results}, we now detail how it could have real-world contributions and practical impact. If deployed in a shopping assistant, our approach could be used to enable automatic question suggestions to be generated and shown to customers across all products in the catalog within a conversational experience. Various configurations of the customer experience can then be evaluated via A/B testing to determine the optimal configuration.

\paragraph{\bf Offline-Online Evaluation Alignment:}
Based on our offline evaluation, we expect a good quality mix of different question types to be also generated online (e.g., generic, recommendations, comparisons, feature based as in Table~\ref{tab:examples}), ranging from short and concise to long and complex. 
However, some generated questions may have stylistic issues, such as the inclusion of first person or second person pronouns. Some generated questions may also be preference elicitation questions that an assistant might ask a customer, rather than a question that a customer might ask an assistant.

\paragraph{\bf Potential Practical Impact:}

As described in Section~\ref{sec:introduction}, incorporating a question suggestion feature within a commercial shopping assistant is a novel and useful real-world application, that millions of customers can adopt and engage with. %
Recommending a diverse mix of both broad and product-specific questions to customers can result in favorable outcomes such as increased engagement time with the assistant and clicks, since it is both easier and faster for customers to click on the suggested questions, rather than type their own. %
Question suggestions followed by their answers %
can impart new and beneficial information to customers about product aspects, that they may not have known or considered during their shopping journey, thus %
increasing customers' satisfaction with the shopping assistant. 
Mechanisms such as caching specific model outputs or processing and delivering the generated tokens in a streaming fashion can effectively reduce the latency of real-time question generation. 
Overall, there is immense potential in using LLM-based question suggestions to enrich the purchase experience of real world customers conversing with shopping assistants.

\section{Conclusion and Future Work}

We focused on improving user adoption of shopping assistants using contextual question suggestions, %
to create a more convenient and seamless experience for customers. 
We proposed a framework %
to automatically generate relevant, useful, answerable and diverse product questions from salient product context, that can be suggested by a shopping assistant to customers as helpful conversation starters, continuations, or inspirations for their own questions.

Looking forward, we envision generative AI to significantly evolve into an even more integral part of the conversational shopping landscape, with digital assistants acting as personalized shopping guides for online buyers. 
In particular, the task of automatically generating beneficial and answerable suggestion questions that a customer may ask a shopping assistant is an essential and challenging problem in the e-commerce domain, to promote faster and smoother conversational interactions. %
There is scope for future improvement, especially to generate more useful and answerable product question suggestions in the desired style. One way of achieving this could be through the use of instruction fine-tuning~\cite{zhang2023instruction} with our labeled data, a direction we leave for future work. 

We also plan to extend our approach to incorporate multi-turn conversational history into the input context. 
We seek to generate personalized, customer-specific product question suggestions based on their past search or purchase behavior.   
We further aim to leverage customer interaction and/or feedback signals (e.g., clicks, likes) to control, modify and improve the questions being generated and presented to customers. 
Such developments will contribute towards boosting the overall dialogue quality, enriching customers' e-commerce experiences, and promoting deeper, enhanced digital interactions that are on-par with human shopping assistants.

\section*{Acknowledgements}
We sincerely thank Eugene Agichtein, Besnik Fetahu, Giuseppe Castellucci, Jason Choi, Zhiyu Chen, Saar Kuzi and Elad Kravi for their valuable feedback and participation in the CBQG project. We are also grateful to the reviewers for their insightful comments.

\section*{Presenter Biography}

Nikhita Vedula is an Applied Scientist at Amazon within the Shopping and Search organization. She received her B.S. degree from the National Institute of Technology, Nagpur, India in 2015, and her Ph.D. from the Ohio State University in 2020. Her research interests and contributions span the fields of natural language processing, conversational AI and information retrieval. She has published more than twenty research papers in leading peer-reviewed conferences such as SIGIR, WSDM, the Web conference, EMNLP and ACL. Her research has been recognized with one Best Paper award, one Best Paper Honorable Mention, and one Outstanding Paper award. She also regularly serves as a Program Committee member and reviewer of several prestigious conferences in her field such as SIGIR, ACL, the Web Conference, WSDM and CIKM.

\bibliographystyle{ACM-Reference-Format}
\balance
\bibliography{sample-base}


\begin{thebibliography}{19}

%%% the Web address.

\ifx \showCODEN    \undefined \def \showCODEN     #1{\unskip}     \fi
\ifx \showDOI      \undefined \def \showDOI       #1{#1}\fi
\ifx \showISBNx    \undefined \def \showISBNx     #1{\unskip}     \fi
\ifx \showISBNxiii \undefined \def \showISBNxiii  #1{\unskip}     \fi
\ifx \showISSN     \undefined \def \showISSN      #1{\unskip}     \fi
\ifx \showLCCN     \undefined \def \showLCCN      #1{\unskip}     \fi
\ifx \shownote     \undefined \def \shownote      #1{#1}          \fi
\ifx \showarticletitle \undefined \def \showarticletitle #1{#1}   \fi
\ifx \showURL      \undefined \def \showURL       {\relax}        \fi
% invisible to TeX
\providecommand\bibfield[2]{#2}
\providecommand\bibinfo[2]{#2}
\providecommand\natexlab[1]{#1}
\providecommand\showeprint[2][]{arXiv:#2}

\bibitem[Achiam et~al\mbox{.}(2023)]%
        {achiam2023gpt}
\bibfield{author}{\bibinfo{person}{Josh Achiam}, \bibinfo{person}{Steven
  Adler}, \bibinfo{person}{Sandhini Agarwal}, \bibinfo{person}{Lama Ahmad},
  \bibinfo{person}{Ilge Akkaya}, \bibinfo{person}{Florencia~Leoni Aleman},
  \bibinfo{person}{Diogo Almeida}, \bibinfo{person}{Janko Altenschmidt},
  \bibinfo{person}{Sam Altman}, \bibinfo{person}{Shyamal Anadkat},
  {et~al\mbox{.}}} \bibinfo{year}{2023}\natexlab{}.
\newblock \showarticletitle{Gpt-4 technical report}.
\newblock \bibinfo{journal}{\emph{arXiv preprint arXiv:2303.08774}}
  (\bibinfo{year}{2023}).
\newblock


\bibitem[Balakrishnan and Dwivedi(2021)]%
        {balakrishnan2021conversational}
\bibfield{author}{\bibinfo{person}{Janarthanan Balakrishnan} {and}
  \bibinfo{person}{Yogesh~K Dwivedi}.} \bibinfo{year}{2021}\natexlab{}.
\newblock \showarticletitle{Conversational commerce: entering the next stage of
  AI-powered digital assistants}.
\newblock \bibinfo{journal}{\emph{Annals of Operations Research}}
  (\bibinfo{year}{2021}), \bibinfo{pages}{1--35}.
\newblock


\bibitem[Bihani et~al\mbox{.}(2018)]%
        {bihani2018faqtor}
\bibfield{author}{\bibinfo{person}{Ankita Bihani}, \bibinfo{person}{Jeffrey~D
  Ullman}, {and} \bibinfo{person}{Andreas Paepcke}.}
  \bibinfo{year}{2018}\natexlab{}.
\newblock \showarticletitle{FAQtor: Automatic FAQ generation using online
  forums}. In \bibinfo{booktitle}{\emph{International Conference on Educational
  Data Mining}}. \bibinfo{pages}{529--532}.
\newblock


\bibitem[Deng et~al\mbox{.}(2023)]%
        {deng2023product}
\bibfield{author}{\bibinfo{person}{Yang Deng}, \bibinfo{person}{Wenxuan Zhang},
  \bibinfo{person}{Qian Yu}, {and} \bibinfo{person}{Wai Lam}.}
  \bibinfo{year}{2023}\natexlab{}.
\newblock \showarticletitle{Product Question Answering in E-Commerce: A
  Survey}.
\newblock \bibinfo{journal}{\emph{arXiv preprint arXiv:2302.08092}}
  (\bibinfo{year}{2023}).
\newblock


\bibitem[Gao et~al\mbox{.}(2023)]%
        {gao2023retrieval}
\bibfield{author}{\bibinfo{person}{Yunfan Gao}, \bibinfo{person}{Yun Xiong},
  \bibinfo{person}{Xinyu Gao}, \bibinfo{person}{Kangxiang Jia},
  \bibinfo{person}{Jinliu Pan}, \bibinfo{person}{Yuxi Bi}, \bibinfo{person}{Yi
  Dai}, \bibinfo{person}{Jiawei Sun}, {and} \bibinfo{person}{Haofen Wang}.}
  \bibinfo{year}{2023}\natexlab{}.
\newblock \showarticletitle{Retrieval-augmented generation for large language
  models: A survey}.
\newblock \bibinfo{journal}{\emph{arXiv preprint arXiv:2312.10997}}
  (\bibinfo{year}{2023}).
\newblock


\bibitem[Ge et~al\mbox{.}(2024)]%
        {ge2024openagi}
\bibfield{author}{\bibinfo{person}{Yingqiang Ge}, \bibinfo{person}{Wenyue Hua},
  \bibinfo{person}{Kai Mei}, \bibinfo{person}{Juntao Tan},
  \bibinfo{person}{Shuyuan Xu}, \bibinfo{person}{Zelong Li},
  \bibinfo{person}{Yongfeng Zhang}, {et~al\mbox{.}}}
  \bibinfo{year}{2024}\natexlab{}.
\newblock \showarticletitle{Openagi: When llm meets domain experts}.
\newblock \bibinfo{journal}{\emph{Advances in Neural Information Processing
  Systems}}  \bibinfo{volume}{36} (\bibinfo{year}{2024}).
\newblock


\bibitem[Ghanem et~al\mbox{.}(2022)]%
        {ghanem-etal-2022-question}
\bibfield{author}{\bibinfo{person}{Bilal Ghanem}, \bibinfo{person}{Lauren
  Lutz~Coleman}, \bibinfo{person}{Julia Rivard~Dexter},
  \bibinfo{person}{Spencer von~der Ohe}, {and} \bibinfo{person}{Alona Fyshe}.}
  \bibinfo{year}{2022}\natexlab{}.
\newblock \showarticletitle{Question Generation for Reading Comprehension
  Assessment by Modeling How and What to Ask}. In
  \bibinfo{booktitle}{\emph{Findings of the Association for Computational
  Linguistics: ACL 2022}}, \bibfield{editor}{\bibinfo{person}{Smaranda
  Muresan}, \bibinfo{person}{Preslav Nakov}, {and} \bibinfo{person}{Aline
  Villavicencio}} (Eds.). \bibinfo{publisher}{Association for Computational
  Linguistics}, \bibinfo{address}{Dublin, Ireland},
  \bibinfo{pages}{2131--2146}.
\newblock
\urldef\tempurl%
\url{https://doi.org/10.18653/v1/2022.findings-acl.168}
\showDOI{\tempurl}


\bibitem[Kojima et~al\mbox{.}(2022)]%
        {kojima2022large}
\bibfield{author}{\bibinfo{person}{Takeshi Kojima},
  \bibinfo{person}{Shixiang~Shane Gu}, \bibinfo{person}{Machel Reid},
  \bibinfo{person}{Yutaka Matsuo}, {and} \bibinfo{person}{Yusuke Iwasawa}.}
  \bibinfo{year}{2022}\natexlab{}.
\newblock \showarticletitle{Large language models are zero-shot reasoners}.
\newblock \bibinfo{journal}{\emph{Advances in neural information processing
  systems}}  \bibinfo{volume}{35} (\bibinfo{year}{2022}),
  \bibinfo{pages}{22199--22213}.
\newblock


\bibitem[Longpre et~al\mbox{.}(2023)]%
        {longpre2023flan}
\bibfield{author}{\bibinfo{person}{Shayne Longpre}, \bibinfo{person}{Le Hou},
  \bibinfo{person}{Tu Vu}, \bibinfo{person}{Albert Webson},
  \bibinfo{person}{Hyung~Won Chung}, \bibinfo{person}{Yi Tay},
  \bibinfo{person}{Denny Zhou}, \bibinfo{person}{Quoc~V Le},
  \bibinfo{person}{Barret Zoph}, \bibinfo{person}{Jason Wei}, {et~al\mbox{.}}}
  \bibinfo{year}{2023}\natexlab{}.
\newblock \showarticletitle{The flan collection: Designing data and methods for
  effective instruction tuning}.
\newblock \bibinfo{journal}{\emph{arXiv preprint arXiv:2301.13688}}
  (\bibinfo{year}{2023}).
\newblock


\bibitem[Mass et~al\mbox{.}(2020)]%
        {mass2020unsupervised}
\bibfield{author}{\bibinfo{person}{Yosi Mass}, \bibinfo{person}{Boaz Carmeli},
  \bibinfo{person}{Haggai Roitman}, {and} \bibinfo{person}{David Konopnicki}.}
  \bibinfo{year}{2020}\natexlab{}.
\newblock \showarticletitle{Unsupervised FAQ retrieval with question generation
  and BERT}. In \bibinfo{booktitle}{\emph{Proceedings of the 58th Annual
  Meeting of the Association for Computational Linguistics}}.
  \bibinfo{pages}{807--812}.
\newblock


\bibitem[Mulla and Gharpure(2023)]%
        {mulla2023automatic}
\bibfield{author}{\bibinfo{person}{Nikahat Mulla} {and} \bibinfo{person}{Prachi
  Gharpure}.} \bibinfo{year}{2023}\natexlab{}.
\newblock \showarticletitle{Automatic question generation: a review of
  methodologies, datasets, evaluation metrics, and applications}.
\newblock \bibinfo{journal}{\emph{Progress in Artificial Intelligence}}
  \bibinfo{volume}{12}, \bibinfo{number}{1} (\bibinfo{year}{2023}),
  \bibinfo{pages}{1--32}.
\newblock


\bibitem[Ni et~al\mbox{.}(2019)]%
        {ni2019justifying}
\bibfield{author}{\bibinfo{person}{Jianmo Ni}, \bibinfo{person}{Jiacheng Li},
  {and} \bibinfo{person}{Julian McAuley}.} \bibinfo{year}{2019}\natexlab{}.
\newblock \showarticletitle{Justifying recommendations using distantly-labeled
  reviews and fine-grained aspects}. In \bibinfo{booktitle}{\emph{Proceedings
  of the 2019 conference on empirical methods in natural language processing
  and the 9th international joint conference on natural language processing
  (EMNLP-IJCNLP)}}. \bibinfo{pages}{188--197}.
\newblock


\bibitem[Rawte et~al\mbox{.}(2023)]%
        {rawte2023survey}
\bibfield{author}{\bibinfo{person}{Vipula Rawte}, \bibinfo{person}{Amit Sheth},
  {and} \bibinfo{person}{Amitava Das}.} \bibinfo{year}{2023}\natexlab{}.
\newblock \showarticletitle{A survey of hallucination in large foundation
  models}.
\newblock \bibinfo{journal}{\emph{arXiv preprint arXiv:2309.05922}}
  (\bibinfo{year}{2023}).
\newblock


\bibitem[Wei et~al\mbox{.}(2022a)]%
        {wei2022emergent}
\bibfield{author}{\bibinfo{person}{Jason Wei}, \bibinfo{person}{Yi Tay},
  \bibinfo{person}{Rishi Bommasani}, \bibinfo{person}{Colin Raffel},
  \bibinfo{person}{Barret Zoph}, \bibinfo{person}{Sebastian Borgeaud},
  \bibinfo{person}{Dani Yogatama}, \bibinfo{person}{Maarten Bosma},
  \bibinfo{person}{Denny Zhou}, \bibinfo{person}{Donald Metzler},
  {et~al\mbox{.}}} \bibinfo{year}{2022}\natexlab{a}.
\newblock \showarticletitle{Emergent abilities of large language models}.
\newblock \bibinfo{journal}{\emph{arXiv preprint arXiv:2206.07682}}
  (\bibinfo{year}{2022}).
\newblock


\bibitem[Wei et~al\mbox{.}(2022b)]%
        {wei2022chain}
\bibfield{author}{\bibinfo{person}{Jason Wei}, \bibinfo{person}{Xuezhi Wang},
  \bibinfo{person}{Dale Schuurmans}, \bibinfo{person}{Maarten Bosma},
  \bibinfo{person}{Fei Xia}, \bibinfo{person}{Ed Chi}, \bibinfo{person}{Quoc~V
  Le}, \bibinfo{person}{Denny Zhou}, {et~al\mbox{.}}}
  \bibinfo{year}{2022}\natexlab{b}.
\newblock \showarticletitle{Chain-of-thought prompting elicits reasoning in
  large language models}.
\newblock \bibinfo{journal}{\emph{Advances in Neural Information Processing
  Systems}}  \bibinfo{volume}{35} (\bibinfo{year}{2022}),
  \bibinfo{pages}{24824--24837}.
\newblock


\bibitem[Xiao et~al\mbox{.}(2023)]%
        {xiao2023efficient}
\bibfield{author}{\bibinfo{person}{Guangxuan Xiao}, \bibinfo{person}{Yuandong
  Tian}, \bibinfo{person}{Beidi Chen}, \bibinfo{person}{Song Han}, {and}
  \bibinfo{person}{Mike Lewis}.} \bibinfo{year}{2023}\natexlab{}.
\newblock \showarticletitle{Efficient streaming language models with attention
  sinks}.
\newblock \bibinfo{journal}{\emph{arXiv preprint arXiv:2309.17453}}
  (\bibinfo{year}{2023}).
\newblock


\bibitem[Zhang et~al\mbox{.}(2023a)]%
        {zhang2023instruction}
\bibfield{author}{\bibinfo{person}{Shengyu Zhang}, \bibinfo{person}{Linfeng
  Dong}, \bibinfo{person}{Xiaoya Li}, \bibinfo{person}{Sen Zhang},
  \bibinfo{person}{Xiaofei Sun}, \bibinfo{person}{Shuhe Wang},
  \bibinfo{person}{Jiwei Li}, \bibinfo{person}{Runyi Hu},
  \bibinfo{person}{Tianwei Zhang}, \bibinfo{person}{Fei Wu}, {et~al\mbox{.}}}
  \bibinfo{year}{2023}\natexlab{a}.
\newblock \showarticletitle{Instruction tuning for large language models: A
  survey}.
\newblock \bibinfo{journal}{\emph{arXiv preprint arXiv:2308.10792}}
  (\bibinfo{year}{2023}).
\newblock


\bibitem[Zhang et~al\mbox{.}(2023b)]%
        {zhang2023siren}
\bibfield{author}{\bibinfo{person}{Yue Zhang}, \bibinfo{person}{Yafu Li},
  \bibinfo{person}{Leyang Cui}, \bibinfo{person}{Deng Cai},
  \bibinfo{person}{Lemao Liu}, \bibinfo{person}{Tingchen Fu},
  \bibinfo{person}{Xinting Huang}, \bibinfo{person}{Enbo Zhao},
  \bibinfo{person}{Yu Zhang}, \bibinfo{person}{Yulong Chen}, {et~al\mbox{.}}}
  \bibinfo{year}{2023}\natexlab{b}.
\newblock \showarticletitle{Siren's song in the AI ocean: a survey on
  hallucination in large language models}.
\newblock \bibinfo{journal}{\emph{arXiv preprint arXiv:2309.01219}}
  (\bibinfo{year}{2023}).
\newblock


\bibitem[Zhuang et~al\mbox{.}(2024)]%
        {zhuang2024toolqa}
\bibfield{author}{\bibinfo{person}{Yuchen Zhuang}, \bibinfo{person}{Yue Yu},
  \bibinfo{person}{Kuan Wang}, \bibinfo{person}{Haotian Sun}, {and}
  \bibinfo{person}{Chao Zhang}.} \bibinfo{year}{2024}\natexlab{}.
\newblock \showarticletitle{Toolqa: A dataset for llm question answering with
  external tools}.
\newblock \bibinfo{journal}{\emph{Advances in Neural Information Processing
  Systems}}  \bibinfo{volume}{36} (\bibinfo{year}{2024}).
\newblock


\end{thebibliography}

\end{document}